\documentclass[11pt,a4paper]{article}
\usepackage[hyperref]{acl2020}
\usepackage{times}
\usepackage{latexsym}
\usepackage{float}
\usepackage{amsfonts}
\usepackage{amsmath}

\usepackage{microtype}
\usepackage{lipsum}

\aclfinalcopy 

\title{Reverse Engineering Configurations of Neural Text Generation Models}

\author{Yi Tay \\
  Google Research \\
  Mountain View \\
  {\tt yitay@google.com} \\
  \And
Dara Bahri \\
  Google Research \\
  Mountain View \\
  {\tt dbahri@google.com} \\
   \And
Che Zheng\\
  Google Research \\
  Mountain View  \\
  {\tt chezheng@google.com} \\
  \AND
 Clifford Brunk\\
  Google Research \\
  Mountain View  \\
  {\tt cliffbrunk@google.com} \\
  \And
 Donald Metzler\\
  Google Research \\
  Mountain View \\
  {\tt metzler@google.com} \\
  \And
 Andrew Tomkins\\
  Google Research \\
  Mountain View\\ 
  {\tt tomkins@google.com} \\
  }

\date{}

\begin{document}
\maketitle
\begin{abstract}
 This paper seeks to develop a deeper understanding of the fundamental properties of neural text generations models. The study of artifacts that emerge in machine generated text as a result of modeling choices is a nascent research area. Previously, the extent and degree to which these artifacts surface in generated text has not been well studied. In the spirit of better understanding generative text models and their artifacts, we propose the new task of distinguishing which of several variants of a given model generated a piece of text, and we conduct an extensive suite of diagnostic tests to observe whether modeling choices (e.g., sampling methods, top-$k$ probabilities, model architectures, etc.) leave detectable artifacts in the text they generate. Our key finding, which is backed by a rigorous set of experiments, is that such artifacts are present and that different modeling choices can be inferred by observing the generated text alone. This suggests that neural text generators may be more sensitive to various modeling choices than previously thought.
\end{abstract}

\section{Introduction}
The task of generating plausible sounding text from large generative neural networks has garnered significant attention recently \cite{zellers2019defending,radford2019language,keskar2019ctrl}. The study of these models has been a keen area of interest for many, resulting in research pertaining to the behavior of generation methods \cite{holtzman2019curious,fan2018hierarchical,gu2017trainable} as well as modeling techniques \cite{radford2019language,welleck2019neural,dai2019transformer,radford2018improving}.

This paper presents a focused empirical study of text generation artifacts, i.e., detectable `signatures' that originate from certain modeling or decoding choices. There is a growing body of research that has focused on discriminating between human and machine generated texts \cite{gehrmann2019gltr,bakhtin2019real,ippolito2019human}. There is also extensive past research on authorship attribution \cite{sanderson2006short,stamatatos2009survey,stamatatos2018overview}, for which it was always assumed that the authors were humans. This work takes a much more fine-grained approach by learning to distinguish between text generated by different machine variants. Do certain modeling choices leave more artifacts than others? In short, given a piece of generated text, can we determine the model configuration that generated this text?

The utility of our study manifests in multiple ways. First, the unraveling of artifacts in generated text enables better understanding of neural text generators, revealing potential fundamental weaknesses in modeling or generation schemes. Our study provides \textit{relative} comparisons of the extent to which artifacts emerge from different modeling choices. Second, this research advances tracking the provenance and origination of machine generated texts, which has a range of useful applications pertaining to online trust and safety, thereby helping to mitigate the overall risk of these models in the wild. To the best of our knowledge, this is the first systematic and fine-grained study of detectable artifacts present in neural generated text.

\paragraph{Our contributions} The overall contributions of this work can be summarized as follows:
\begin{itemize}
    \item We present a large\-scale analysis of generated text with a special focus on studying artifacts produced by large generative models.
    \item We propose the new task of distinguishing between different fine-grained configurations based on the generated text alone. The key idea is that classifiers performing better than random can capture configuration\-specific artifacts.
    \item Our findings show that (1) modeling choices can be captured by simple classifiers through artifacts that are present in generated text alone, (2) the ease of prediction varies across different hyperparameter configurations, (3) word order is not that important in unraveling artifacts, i.e., artifacts are probably more related to word choice than syntax and composition and (4) distinguishing between model variants is much harder than predicting between \textit{human-or-machine} only.
\end{itemize}

\section{Related Work}
There are many research efforts related to machine generated text. The work in this area can be characterized into two broad categories - (1) learning to generate better text and (2) learning to mitigate against generated text. 

In the former, large generative models such as GPT/GPT-2 \cite{radford2018improving,radford2019language}, CTRL \cite{keskar2019ctrl} and Grover \cite{welleck2019neural} have recently demonstrated the possibility of generating high quality text. The study of sampling methods for auto-regressive models has also been active where sampling methods such as top-$k$ \cite{fan2018hierarchical} and nucleus sampling \cite{holtzman2019curious} have been proposed.

Likewise, there have also been recent ongoing efforts that are targeted at distinguishing human text from machine generated text. \cite{gehrmann2019gltr} proposed GLTR, a visual and statistical tool for aiding the detection of machine generated text. In a similar vein, \cite{bakhtin2019real} proposed energy-based models. Statistical detection of machine generated text is possible largely due to the  the presence of artifacts. To this end, the race between generators and discriminators is not entirely de-coupled. \cite{welleck2019neural} showed that a good generator is also a good discriminator. 

Concurrent work \cite{ippolito2019human} investigates the performance of human raters on the task of detecting machine generated text. Similarly, they also investigate the effect of model hyperparameters with respect to the ease of being detected by human raters.

Our work is also related to the field of authorship attribution \cite{stamatatos2009survey} which tries to identify the author behind a piece of text. A series of shared tasks have been proposed over the years \cite{stamatatos2018overview,tschuggnall2017overview}. The tasks have primarily focused on stylometry and text-based forensics. A key assumption is that authors leave behind distinguishable signatures (or artifacts) in their writings. Along a similar vein, our work re-imagines this task by considering different instances of generative models as authors.

The emergence of artifacts left behind by machine generated text is a peculiar and interesting phenomena. This work takes this direction further by studying the fine-grained artifacts produced by different modeling choices in hopes of better understanding machine generation in general.

\section{Methodology}
In this section, we introduce our experimental settings and setup.
\subsection{Generative Model Configuration}
Our experiments employ Grover \cite{zellers2019defending} as the text generator. We consider three generation configurations in our experiments. They are described as follows:

\begin{itemize}
    \item \textbf{Model Sizes} - Generative models often come with pre-defined sizes that refer to the layer widths and parameterization. For Grover, the model size options include \textit{Base}, \textit{Large}, and \textit{Mega}.
    \item \textbf{Sampling Method} - The sampling function controls the decoding process used to generate text. We explore variants of top-$k$ \cite{fan2018hierarchical}, top-$p$ nucleus sampling \cite{holtzman2019curious}, and associated $p/k$ values.
    \item \textbf{Conditioning} - Length of initial article conditioning. We define $\ell$ which is the amount of text given to the model. The initial $\ell$ tokens is concatenated at the end of the title sequence for the model to start generating.
\end{itemize}
In the design of our experiments, while there are countless possibilities to search for, we deliberately sought out settings that are most general and/or are considered fine-grained subtle changes. Such subtle changes are likely to be more challenging to detect compared to larger changes. For example, predicting Grover parameterization subsumes the task of distinguishing Grover versus GPT-2. We assume that if a model is able to solve the former, the latter becomes relatively trivial.

\subsection{Classifier Models}
We train a classifier model to discriminate between different model configurations. Generally, the task is framed as a multi-class classification problem where each model configuration is a class that is predicted. Models accept a sequence of tokens as an input. Sequences pass through a parameterized or non-parameterized encoder which are finally passed as input to a softmax classification layer.

In this work, we explore and benchmark the effectiveness of various encoding inductive biases such as recurrent, convolutional, and self-attention based models. This is primarily motivated as a probe into the problem domain, i.e., by witnessing the behaviour of different encoder architectures, we may learn more about the nature of these tasks/datasets.

\paragraph{Inductive Biases} We consider the following encoding architectures (1) \textbf{BoW (Linear)} - a simple bag-of-words (BoW) baseline that averages the word embeddings and passes the average representation into a single linear classifier. $Y=\text{Softmax}(W(X))$. (2) \textbf{BoW (MLP)} - another simple baseline that builds on top of the Linear baseline. We add a single nonlinear layer with ReLU activation function, i.e., $Y=\text{Softmax}(W_{2}\sigma_{r}(W_{1}(X)))$. (3) \textbf{ConvNet} - We consider a 1D Convolution layer of filter width $3$. We convolve over the input embeddings and pass the average (representation) into a linear Softmax classification layer. (4) \textbf{LSTM} - Similar to the CNN model, we encode the input sequence with an LSTM layer and pass the mean-pooled representation into a Softmax layer. (4) \textbf{Transformer Encoders} - We use 4-layered multi-headed Transformer \cite{vaswani2017attention} encoders with multi-head self-attention. 


\begin{table}[ht]
\centering
\small
\begin{tabular}{c|c}
\hline
Task Name & Classes \\ 
\hline
$p$-Samp (P1) &  $p \in [0.95,0.90,0.85]$\\
$p$-Samp (P2) &  $p \in [0.95,0.85,0.75]$\\
$p$-Samp (P3) & $p \in [0.95,0.90,0.85,0.80,0.75]$\\
\hline 
$k$-Samp (K1) &  $k \in [10,20,30]$\\
$k$-Samp (K2) &  $k \in [10,30,50]$\\
$k$-Samp (K3) & $k \in [10,20,30,40,50]$\\
\hline
Cond (C1) & $ \ell \in [10,50,100]$ \\
Cond (C2) & $ \ell \in [10,20,30]$ \\
Cond (C3) & $ \ell \in [10,20,30,40,50]$ \\
\hline
Size (S1) & $S \in \{Base,Large,Mega\}$\\
 \hline
\end{tabular}
\caption{List of proposed Machine Configuration Discrimination (MCD) tasks.}
\label{mad_tasks}
\end{table}

\begin{table*}[h]


\begin{center}
\small
\begin{tabular}{ c|cccccccccc|c} 
 \hline
 Model/Task & P1 &	P2 & P3 &	K1 &	K2  &	K3 &	C1 &	C2 &C3 &	S1 & AVG \\
 \hline
Chance	& 33.33	& 33.33	& 20.00	& 33.33	& 33.33	& 20.00	& 33.33& 	33.33& 	20.00& 	33.33& 	29.33\\
BoW Linear & 	55.22& 	69.17& 	55.88& 	54.50& 	62.82& 	38.39& 	42.27& 	34.66& 	21.96& 	43.70	& 47.86\\ 
BoW MLP	& 55.19	& 69.71	& 56.87& 	56.11& 	62.65& 	39.97& 	42.89& 	34.62& 	22.70& 	43.21& 	48.39\\
CNN	& 55.37	& 69.65	& 57.47	& 55.54& 	63.94& 	40.30& 	43.04& 	35.06& 	23.10& 	43.74& 	48.72\\
LSTM& 	54.89& 	68.88& 	54.52	& 54.97& 	62.71& 	40.19& 	45.69& 	34.00& 	23.76	& 43.51& 	48.31\\
Transformer	& 53.74& 	70.23& 	59.73& 	55.24& 	63.40& 	40.48& 	43.94& 	34.35& 	24.00& 	42.24& 	48.74\\
\hline
\% Gain & +66$\%$ & +111$\%$ & +199\% & +68\% & +92\%& +21\% & +37\% & +5\% & +20\% & +31\% & +66\% \\
 \hline 
\end{tabular}
\caption{Results on machine configuration detection. $\%$ gain provides a general sense of how prevalent artifacts are for a given configuration.}
\label{table1}
\end{center}
\begin{center}
\small
\begin{tabular}{ c|cccccccccc|c} 
 \hline
 Model/Task & P1$_h$ &	P2$_h$ & P3$_h$ &	K1$_h$ &	K$_h$  &	K3$_h$ &	C1$_h$ &	C2$_h$ &C3$_h$ &	S1$_h$ & AVG \\
 \hline
Chance &	25.00&	25.00&	16.67&	25.00	&25.00&	16.67&	25.00 &	25.00&	33.33&	25.00&	24.17\\
BoW Linear &	67.49&	76.60&	63.81&	73.27	&78.90	&57.47	&47.30&	46.10&	33.19&	58.58	&60.27\\
BoW MLP &	67.98&	76.73&	65.58&	74.06&	78.87&	57.22&	49.15&	47.45&	33.85&	58.24&	60.91\\
CNN&	68.38&	75.61&	64.79&	73.25&	78.76&	57.15	&49.35&	47.46&	33.88&	58.59&	60.72\\
LSTM&	69.03&	77.04&	68.69&	74.36&	78.64&	57.90&	50.45&	48.35&	34.33&	58.10&	61.69\\
Transformer &	68.99&	78.63&	68.58&	74.56&	79.26&	57.17&	50.92&	48.65&	35.23&	59.63&	62.16\\
 \hline 
 \% Gain & +$176\%$ & +$215\%$ & +312\% & +198\% & +217\%& +247\% & +104\% & +95\% & +6\% & +139\% & +157\% \\
 \hline 
\end{tabular}
\caption{Results on the machine configuration detection tasks with human articles as an additional class.}
\label{h2}
\end{center}
\end{table*}

\subsection{Experimental Setup} 
This section outlines our experimental setup.
\paragraph{News Corpora} As a seed corpus, we use the CNN/Dailymail news corpus. This corpus is widely used in other NLP tasks \cite{hermann2015teaching} such as question answering and summarization. The CNN/Dailymail corpus comprises approximately $90K$ news articles. Given an initial seed corpora of $N$ news articles, we generate an additional collection of $N$ machine generated articles for each configuration. 

\paragraph{Tasks} We define \textbf{ten} tasks as described in Table \ref{mad_tasks}. These tasks aim at predicting the correct model configuration given the generated text. For all tasks, we use a maximum sequence length of $500$ and split the dataset into 80\%/10\%/10\% train, development, and testing splits. We include an additional variant $+h$ which denotes that we add the human\-written article as an additional class to the mix.

\paragraph{Model Training} For all models, we fix the word embeddings to $d=64$. Embeddings are trained from scratch. All encoder hidden unit size is also set to $64$. We tuned the dimensions of models in the range of $d \in \{16,32,64,128,256\}$ and found no noticable improvement beyond $d=64$. We train all models for $50$ epochs with a batch size of $64$. We employ early stopping with patience 3 if validation accuracy does not improve. Final test accuracy is reported based on the best results on the validation set.

\section{Insights and Findings}
This section presents the insights and findings uncovered by our experiments. Table \ref{table1} and Table \ref{h2} present the core of our experimental results.
\paragraph{(1) Artifacts are found.} 
Our experiments show that simple classifiers are able to distinguish fine-grained and subtle differences between modeling choices (e.g., top-$p$ probabilities or condition length $\ell$) in generated texts. In Table \ref{table1}, we observe that all classifiers have an accuracy much higher than random chance (almost \textbf{double} in some cases), which suggests that distinguishing between different classes is relatively straightforward. In short, we are able to empirically conclude that all modeling choices leave behind some form of detectable artifacts.
\paragraph{(2) Different generating choices leave behind different amounts of artifacts.} From Table \ref{table1}, the difficulty of each task generally depends on the specific modeling choice. For example, distinguishing between model size (S1) is much harder than the top-$p$ value. Overall, we observe that methods that directly operate at the generation level (sampling $p$ or $k$ values) are much easier to predict (i.e., leave more artifacts) than condition length ($C1,C2$) or model size ($S1$). It is a somewhat surprising result that varying the initial condition length leaves artifacts in the generated text.

A secondary finding is that discriminating $p$ or $k$ values that are close together is a significantly more challenging task than those that are far apart (i.e., task P1 vs P2). This empirically shows that generated text moves along some form of ordering and magnitude, i.e., $s(a,b) \leq s(b,c)$ if $a-b>b-c$ where $a,b,c \in \mathbb{R}$ and $s(x,y)$ is the accuracy score obtained by classifying between configurations $x,y$. 
\paragraph{(3) Word order does not matter too much.} The key observation when pitting various sequence encoding inductive biases against each other is to observe if modeling sequential interactions (short-term or long-range dependencies) and/or word order helps in any of the MCD tasks. The observation is that most complex encoders that takes into account word order do not outperform simple BoW (bag of words) with linear classifiers. This suggests that artifacts found in the text are mostly related to style (e.g., word choices), as opposed to compositional dependencies (e.g., word order). Occasionally, we observe some marginal gains when utilizing ConvNet or Transformers. We hypothesize that considering some amount of token interaction is indeed useful, albeit very marginally. Moreover, the recurrent model (LSTM) performs worse in most cases, suggesting that complex compositional relations are not necessary to capture artifacts.

\paragraph{(4) Discriminating between machines is harder than human and machine.}
Table \ref{h2} report the results of MCD tasks with an additional human article class. By adding human generated articles into the mix, the classification accuracy increases ($\approx10\%$) across all tasks. Upon inspection, we find that the model separates the human written articles at beyond $90\%$ accuracy, which leads to an overall increase in performance. Hence, the task of distinguishing between machine-machine text is much harder than distinguishing between human-machine text.

\section{Discussion}
This section discusses the implications of our results and findings.
\paragraph{(1) The sensitivity of neural text generation models emerge as artifacts in the generated text.} Our results show that a state-of-the-art text generation model produces significant amounts of artifacts even when making small hyperparameter changes (such as sampling probabilities). It is also relatively surprising that the amount of article conditioning and model size can also be predicted to a certain degree. We feel that this might arise from limitations in the design of neural generation models which may warrant further study.
\paragraph{(2) Tracing the provenance and origination of text generation models is easier than expected.} Given that minor changes to decoding settings leave distinguishable signatures, we hypothesize that it is relatively easy to trace and cluster content produced by specific generative models.

\section{Conclusion}
We studied machine generated text and found that modeling choices leave artifacts, i.e., it is possible to predict modeling choices such as parameterization/sampling choices by looking at generated text alone. We proposed the novel task of machine configuration detection (MCD) which aided in the discovery of these artifacts. We believe our work paves the way for better understanding of neural text generation models and understanding that modeling choices reveals the model configurations is a first crucial step.

\bibliography{acl2020}
\bibliographystyle{acl_natbib}

\appendix
\end{document}